\documentclass[preprint,12pt]{elsarticle}
\usepackage{graphicx}
\usepackage{multirow}
\usepackage[table]{xcolor}
\usepackage[table,dvipsnames]{xcolor}
\usepackage{hyperref}
\usepackage{array}
\usepackage{array}
\newcolumntype{C}{>{\centering\arraybackslash}p{1.8cm}}
\usepackage{amssymb}
\usepackage{amsmath}
\usepackage{multirow}
\usepackage{pifont}
\usepackage[numbers]{natbib}
\usepackage{booktabs}
\usepackage{tabularx}
\usepackage{algorithm}
\usepackage{algpseudocode}
\usepackage{placeins}

\newcommand{\cmark}{\ding{51}}
\newcommand{\xmark}{\ding{55}}

\hypersetup{
  colorlinks=true,
  citecolor=blue,
  linkcolor=blue,
  urlcolor=blue
}

\renewcommand{\arraystretch}{1.15}
\journal{Nuclear Physics B}

\begin{document}
\begin{frontmatter}
\title{Normality-Preserving Continual Industrial Anomaly Detection via Orthogonal LoRA Banks}

\author[ee]{Weibai Fang}
\author[ee]{Haijun Che\corref{cor1}}
\ead{hjche@ysu.edu.cn}

\author[se]{Feiyang Ren}
\author[sc]{Qiancheng Lao}

\cortext[cor1]{Corresponding author}

\affiliation[ee]{organization={School of Electrical Engineering, Yanshan University},
            addressline={Haigang District},
            city={Qinhuangdao},
            postcode={066000},
            state={Hebei},
            country={China}}

\affiliation[se]{organization={School of Artificial Intelligence, Yanshan University},
            addressline={Haigang District},
            city={Qinhuangdao},
            postcode={066000},
            state={Hebei},
            country={China}}

\affiliation[sc]{organization={Silesian College of Intelligent Science and Engineering, Yanshan University},
            addressline={Haigang District},
            city={Qinhuangdao},
            postcode={066000},
            state={Hebei},
            country={China}}

\begin{graphicalabstract}
\includegraphics[width=\textwidth]{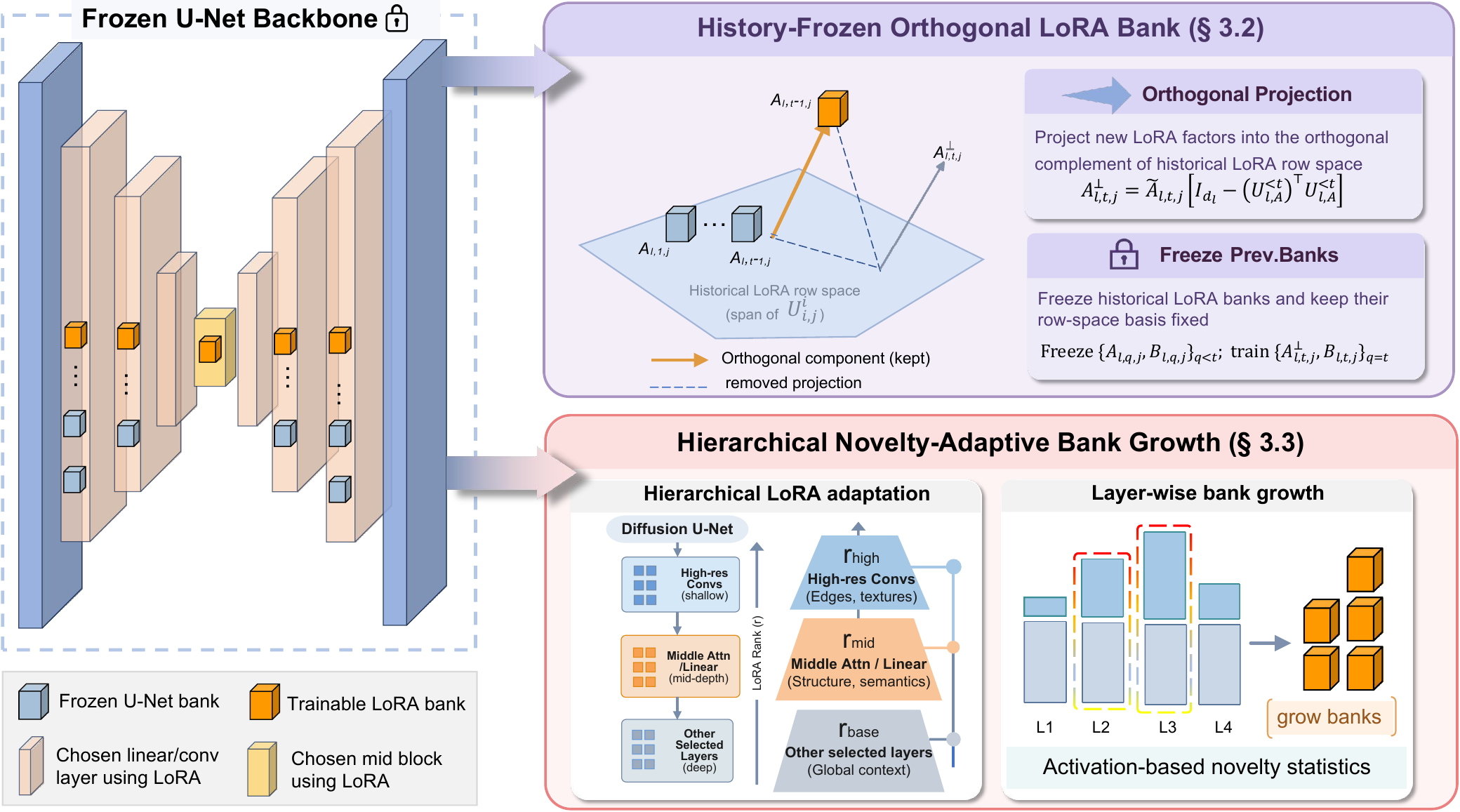}
\end{graphicalabstract}
\begin{highlights}
\item We introduce LoRA into continual diffusion-based IAD and propose HF-OLB, which treats LoRA banks as low-rank normality memory units and orthogonalizes new adapters against frozen historical banks to preserve category-specific normality priors.
\item We propose HNABG, a hierarchical bank-growth strategy that allocates residual normality capacity according to layer-wise subspace novelty, enabling new-category normality acquisition only where historical bases are insufficient.
\item Extensive experiments on continual settings derived from MVTec and VisA demonstrate that the proposed method could
provide stronger normality-prior preservation, lower forgetting, and more stable anomaly localization under long sequential category updates.
\end{highlights}

\begin{abstract}
Continual industrial anomaly detection with diffusion models suffers from historical normality prior drift and catastrophic forgetting. Existing continual diffusion methods preserve previous knowledge through replay or constrained optimization, but they lack an explicit mechanism for isolating and protecting category specific normality priors during sequential adaptation. Although low-rank adaptation provides modular residual updates, standard LoRA neither freezes historical normality subspaces nor prevents new adapters from interfering with previous ones. To address this issue, we propose a normality-preserving continual anomaly detection framework based on two modules: History Frozen Orthogonal LoRA Bank (HF-OLB) and Hierarchical Novelty Adaptive Bank Growth module (HNABG). HF-OLB freezes both the pretrained U Net backbone and the learned LoRA banks, and constrains new task specific normality residuals to the orthogonal complement of historical LoRA subspaces. HNABG further allocates layer dependent residual capacity and expands the bank only when the residual normality novelty exceeds the expressive capacity of existing banks. Extensive experiments on MVTec and VisA demonstrate the effectiveness of the proposed method. On the challenging VisA $2 \times 6$ setting, our method achieves 83.6/91.8 image and pixel level A-AUROC with 3.8/3.9 FM, improving pixel level A-AUROC over the state of the art by 3.2 points while reducing pixel level FM by 1.3. These results show that our method effectively preserves historical normality priors in long horizon continual category sequences.
\end{abstract}


\end{frontmatter}

\section{Introduction}

Industrial anomaly detection (IAD) aims to identify and localize defects using only anomaly-free training samples~\cite{roth2022towards,gudovskiy2022cflow,deng2022anomaly,zavrtanik2021draem,jeong2023winclip,li2024promptad,tang2026enhancing,cao2026hypcv,jia2026aquada,xin2026time,wang2026open}. 
Recent methods have achieved strong performance in static or offline settings, where all categories are assumed to be available during training. 
However, this assumption is often violated in practical production environments, where new product categories are introduced sequentially and the detector must remain effective on previously learned categories. 
This setting raises a central challenge for continual IAD: how to acquire the normal patterns of new categories without drifting away from the normality priors established for historical categories.

Diffusion models are particularly attractive for IAD because their detection ability is closely tied to anomaly-free normality priors. 
A diffusion based detector learns the normal visual manifold from anomaly-free samples and reconstructs anomalous regions toward their corresponding normal appearances. 
The discrepancy between the input image and its reconstructed normal counterpart can then be used to localize defects~\cite{rombach2022high,zhang2023unsupervised,lu2023removing,yao2024glad,fuvcka2024transfusion,he2024diffusion,yan2024anomalysd,sakai2025invad}. 
Beyond industrial inspection, diffusion models have also shown strong controllable generation and editing capabilities in subject transformation, layout consistent editing, and fine grained garment generation~\cite{shen2025imagedit,shen2025imagharmony,shen2025imaggarment}, further demonstrating their potential as expressive visual priors. 
Nevertheless, when diffusion based IAD is deployed in a continual category update setting, adaptation to new categories may perturb the normality priors learned for previous categories. 
Once such historical priors drift, the model may fail to reconstruct old category anomalies into faithful normal appearances, leading to degraded anomaly localization and catastrophic forgetting.

General continual learning methods preserve previous knowledge through replay, regularization, parameter isolation, or constrained optimization~\cite{kirkpatrick2017overcoming,wang2022continual,tang2024incremental,liu2024unsupervised}. 
Although these strategies reduce interference between old and new tasks in a general representation space, they do not explicitly model the category specific normality priors required by diffusion based IAD. 
This limitation is critical for reconstruction based anomaly localization, since historical normality priors directly determine whether anomalies from old categories can still be restored into normal appearances. 
Recent continual diffusion based IAD methods, such as CDAD~\cite{li2025one} and ReplayCAD~\cite{hu2025replaycad}, attempt to address sequential category updates under the diffusion framework. 
However, they mainly preserve previous knowledge at the model or sample level, rather than explicitly isolating historical normality subspaces.

\begin{figure}[t]
    \centering
    \includegraphics[width=0.95\textwidth]{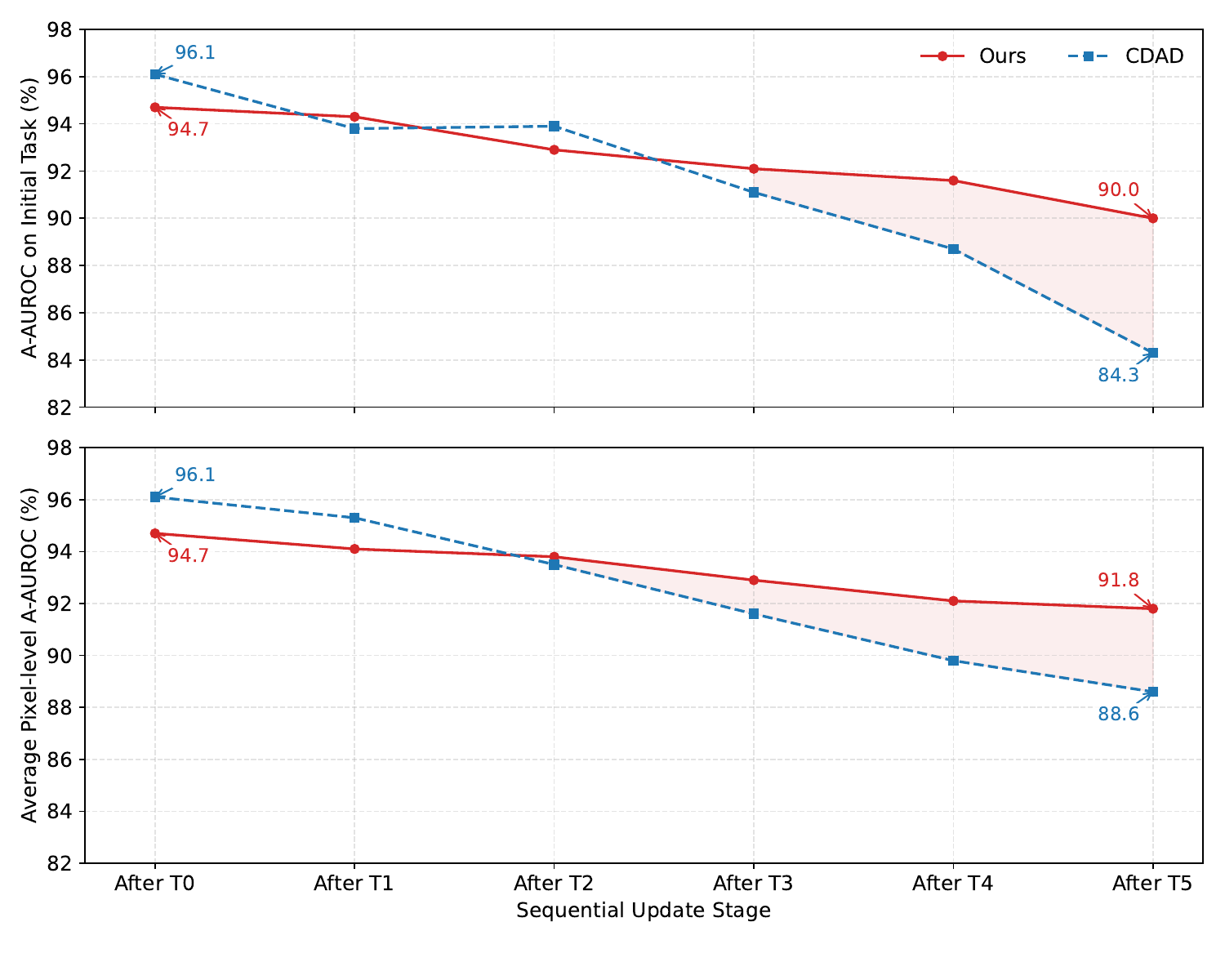}
    \caption{Normality prior retention on VisA setting 3 under sequential category updates. 
    The upper panel reports the A-AUROC on the initial task after each update, where the initial task contains the first two categories in the continual stream and is re-evaluated after subsequent tasks arrive. 
    The lower panel reports the average pixel level A-AUROC over all seen tasks after each update. 
    The slower degradation of our method indicates better preservation of historical normality priors.}
    \label{fig:introduction}
\end{figure}

As shown in Figure~\ref{fig:introduction}, CDAD suffers from a clear performance decline as new categories arrive, indicating that general knowledge retention does not necessarily prevent historical normality prior drift. 
Therefore, continual diffusion based IAD requires a dedicated mechanism that preserves historical normality priors while incorporating residual normality information from newly arriving categories.

Low Rank Adaptation (LoRA) is naturally suited to continual learning because it separates task induced residual updates from the frozen backbone and provides modular carriers for newly acquired knowledge~\cite{hu2021lora,oh2025aulora,yang2025uniad,liang2024inflora,he2025cllora,wei2025onlinelora}. 
In diffusion based IAD, we reinterpret each LoRA bank as a low rank normality memory unit that stores category specific normality residuals on top of a frozen diffusion U Net. 
However, standard LoRA does not explicitly protect historical normality subspaces. 
Newly trained adapters may overlap with or distort previous LoRA subspaces, causing interference between old and new normality representations.

To address these issues, we propose a continual industrial anomaly detection framework based on a frozen Stable Diffusion U Net. 
The first component, History Frozen Orthogonal LoRA Bank (HF-OLB), serves as a historical normality prior protection mechanism. 
It freezes the LoRA banks learned from previous tasks as historical normality bases and projects newly learned adapters into the orthogonal complement of the historical LoRA subspace. 
By preventing new normality residuals from being written into previously occupied subspaces, HF-OLB reduces interference between old and new normality representations and preserves the reconstruction behavior required by historical categories. 
The second component, Hierarchical Novelty Adaptive Bank Growth (HNABG), allocates residual normality capacity in a layer aware manner. 
Instead of expanding LoRA banks uniformly across all U Net layers, HNABG estimates layer wise subspace novelty and identifies where historical LoRA subspaces are insufficient to express the normal patterns of the incoming category. 
New residual normality capacity is therefore introduced only in layers that contain genuinely novel normality information. 
This selective allocation avoids both insufficient adaptation in critical layers and redundant residual subspaces that may increase future interference.

Extensive experiments on continual settings derived from MVTec Anomaly Detection (MVTec)~\cite{bergmann2019mvtec} and Visual Anomaly (VisA)~\cite{zou2022spot} demonstrate that the proposed method better preserves historical normality priors under sequential category updates. 
On VisA setting 3, after the final update stage, our method achieves 91.8\% average pixel level A-AUROC over all seen tasks, outperforming CDAD by 3.2 percentage points. 
These results show that preserving historical normality priors while adaptively allocating residual normality capacity leads to more stable anomaly localization under long continual category sequences.

The main contributions of this work are threefold:
\begin{itemize}
    \item We introduce LoRA into continual diffusion based IAD and propose HF-OLB, which treats LoRA banks as low rank normality memory units and orthogonalizes new adapters against frozen historical banks to preserve category specific normality priors.

    \item We propose HNABG, a hierarchical bank growth strategy that allocates residual normality capacity according to layer wise subspace novelty, enabling new category normality acquisition only where historical bases are insufficient.

    \item Extensive experiments on continual settings derived from MVTec and VisA demonstrate that the proposed method provides stronger normality prior preservation, lower forgetting, and more stable anomaly localization under long sequential category updates.
\end{itemize}

\section{Related Work}

\subsection{Diffusion Based Anomaly Detection}
\label{subsec1}

Industrial anomaly detection has been widely studied through feature deviation modeling, retrieval, distillation, and vision language priors~\cite{roth2022towards,gudovskiy2022cflow,deng2022anomaly,jeong2023winclip,li2024promptad,zhu2024toward}. 
Different from these paradigms, diffusion based methods model anomaly free visual content and localize defects through reconstruction discrepancies, making them suitable for subtle and fine grained industrial defects.
Early reconstruction based methods, such as DRAEM~\cite{zavrtanik2021draem}, combine image restoration with discriminative segmentation. 
Recent diffusion based methods further improve normal reconstruction by denoising anomalous inputs toward the normal manifold. 
DiffAD~\cite{zhang2023unsupervised} and RANO~\cite{lu2023removing} formulate anomaly removal as a diffusion denoising process. 
GLAD~\cite{yao2024glad} introduces adaptive denoising and spatial fusion, while TransFusion~\cite{fuvcka2024transfusion} integrates reconstruction and localization in an iterative framework. 
RealNet~\cite{zhang2024realnet} improves realistic anomaly synthesis, and AnomalySD~\cite{yan2024anomalysd} adapts Stable Diffusion for few shot multiclass anomaly detection on MVTec~\cite{bergmann2019mvtec} and VisA~\cite{zou2022spot}. 
Inversion based methods such as InvAD~\cite{sakai2025invad} improve efficiency, but may weaken dense spatial restoration, which is important for pixel level localization.
Beyond anomaly detection, diffusion models have shown strong conditional generation ability in pose guided synthesis, talking face generation, story visualization, and virtual dressing~\cite{shen2024advancing,shen2024imagpose,shen2025long,shen2025boosting,shen2025imagdressing}. 
These studies further demonstrate the effectiveness of diffusion priors for controllable visual reconstruction and generation. 
However, most diffusion based IAD methods are designed for static category sets. 
In real production lines, new product categories may arrive sequentially, requiring the detector to update continually while preserving historical normality priors. 
Existing static diffusion based methods do not explicitly separate or protect category specific normality patterns, limiting their reliability in continual IAD.

\subsection{Continual Anomaly Detection}
\label{subsec2}
Continual learning methods aim to reduce forgetting through replay, regularization, parameter isolation, or constrained optimization~\cite{kirkpatrick2017overcoming,wang2022continual}.
In IAD, IUF~\cite{tang2024incremental} introduces an incremental framework for small defect inspection, while UCAD~\cite{liu2024unsupervised} uses continual prompting with a key prompt knowledge memory bank. 
These methods move IAD toward realistic sequential scenarios, but mainly operate in feature space or rely on lightweight reconstruction, offering limited support for preserving pixel level normality priors.
Recent works combine continual learning with diffusion models. 
ReplayCAD~\cite{hu2025replaycad} uses diffusion based generative replay to preserve historical details. 
CDAD~\cite{li2025one} stabilizes diffusion training with gradient projection, iterative singular value decomposition, and anomaly masked conditioning. 
Although these methods improve continual diffusion based IAD, they mainly preserve knowledge through replay samples or constrained model updates. 
They do not explicitly isolate the category specific normality subspaces formed by previous tasks. 
As a result, new category adaptation may still perturb historical normality priors and degrade reconstruction behavior for old categories.

\subsection{LoRA Based Modular Adaptation and Subspace Preservation}
\label{subsec3}
LoRA~\cite{hu2021lora} encodes task induced residual updates on top of a frozen backbone, making it suitable for modular and parameter efficient adaptation. 
In anomaly detection, LoRA has mainly been used for vision language adaptation. 
AULoRA~\cite{oh2025aulora} injects LoRA into CLIP for zero shot anomaly detection, and UniCLIP AD~\cite{yang2025uniad} adapts CLIP for unified cross category industrial inspection. 
These methods show the potential of low rank adaptation, but they are not designed for diffusion reconstruction or continual normality preservation.
LoRA has also been explored in continual learning. 
InfLoRA~\cite{liang2024inflora} constrains new updates to an interference free subspace, CL LoRA~\cite{he2025cllora} separates shared and task specific adapters, and Online LoRA~\cite{wei2025onlinelora} handles task free online continual learning through distribution shift detection and knowledge consolidation. 
However, these methods mainly target recognition or classification, where the goal is to preserve discriminative representations. 
Continual diffusion based IAD has a different requirement: historical normality priors must remain stable so that anomalies from old categories can still be reconstructed into normal appearances.

Overall, existing diffusion based IAD methods model defect free visual content but rarely address continual category updates. 
Continual anomaly detection methods reduce forgetting but do not explicitly protect historical normality subspaces. 
LoRA based continual methods provide modular residual adaptation, but are not tailored to reconstruction based normality preservation. 
These gaps motivate our normality preserving continual diffusion framework, where historical normality priors are stored in frozen LoRA banks and new residual normality information is incorporated through orthogonal and novelty adaptive bank updates.

\section{Method}

\begin{figure}[t]
    \centering
    \includegraphics[width=1\textwidth]{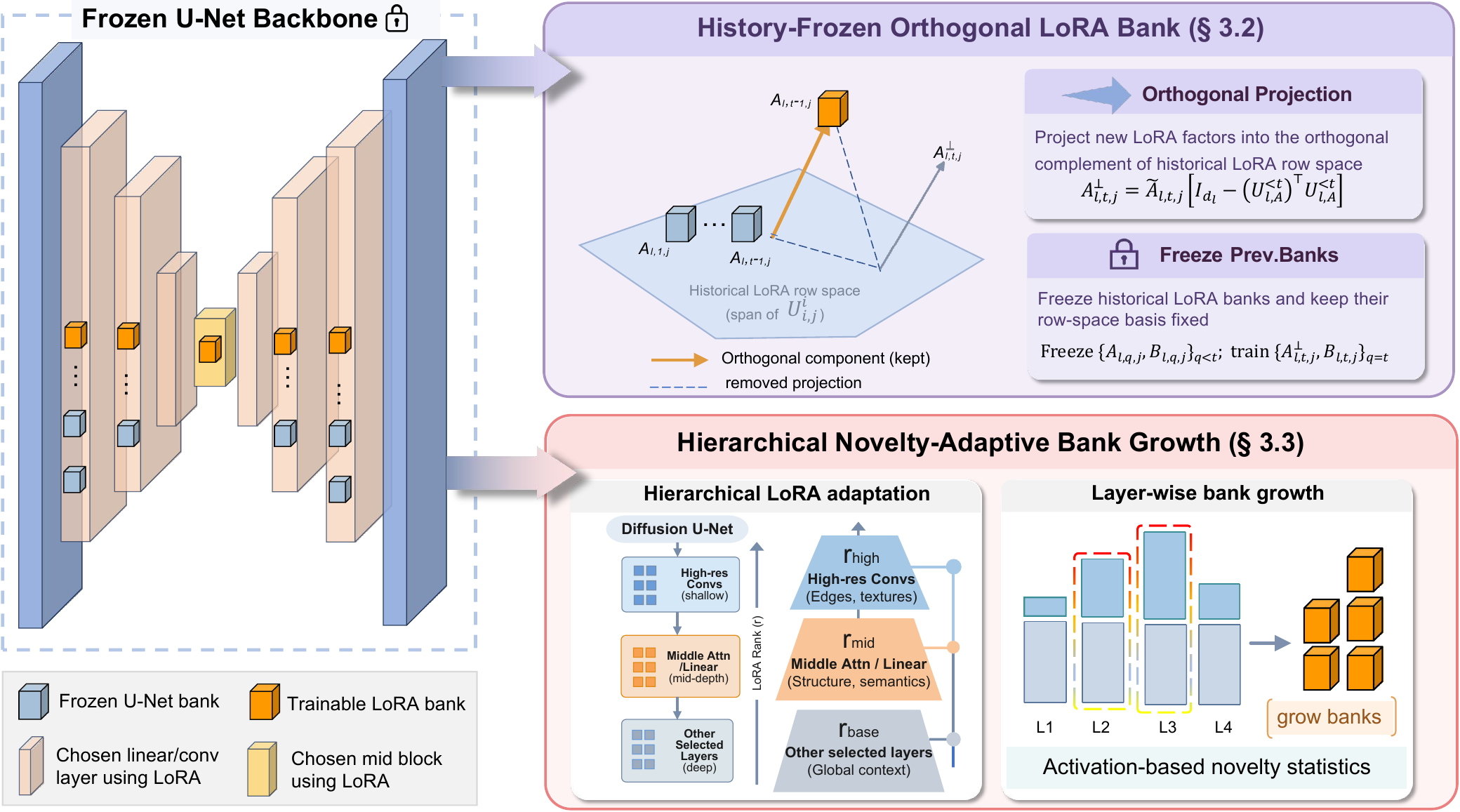}
    \caption{Illustration of our method. The left part shows the overall framework of the continual diffusion model using LoRA. The right part shows the proposed History-Frozen Orthogonal LoRA Bank and Hierarchical Novelty-Adaptive Bank Growth mechanism.}
    \label{fig:overview}
\end{figure}

Figure \ref{fig:overview} provides an overview of our method. The framework consists of three main modules: a frozen U-Net backbone, a history-frozen orthogonal LoRA bank, and a hierarchical novelty-adaptive bank growth model. The frozen U-Net backbone provides a stable diffusion prior for anomaly-free reconstruction, while the history-frozen orthogonal LoRA bank serves as a modular memory for category-specific normality priors. Specifically, previously learned LoRA banks are frozen as historical normality bases, and newly added adapters are projected into the orthogonal complement of the historical subspace, thereby reducing interference between old and new normality representations. The hierarchical novelty-adaptive bank growth model further estimates layer-wise subspace novelty and expands the LoRA bank only for layers where incoming normal patterns cannot be sufficiently represented by historical LoRA subspaces. In this way, the proposed framework preserves historical normality priors while allocating residual normality capacity for newly arriving categories under continual industrial anomaly detection.

\subsection{Stable-Diffusion Reconstruction Backbone}

We build our anomaly detector on a latent diffusion backbone initialized from Stable Diffusion. After adaptation on normal data, the model learns a strong prior of anomaly-free appearance and tends to reconstruct abnormal inputs toward the normal manifold. As a result, abnormal regions are poorly preserved in the reconstruction and become salient in the reconstruction discrepancy map. This property is especially attractive for industrial anomaly detection, where the notion of ``normality'' is often easier to model than the potentially unbounded space of anomalies.

Given an input image $x$, a variational autoencoder first maps it into the latent space,
\begin{equation}
z_0 = E(x),
\end{equation}
where $E(\cdot)$ denotes the VAE encoder. Following the standard diffusion formulation, a noisy latent at denoising timestep $s$ is constructed as
\begin{equation}
z_s = \sqrt{\bar{\alpha}_s} z_0 + \sqrt{1 - \bar{\alpha}_s}\epsilon,
\quad
\epsilon \sim \mathcal{N}(0, I).
\end{equation}
Here, $s$ is used exclusively for the diffusion denoising timestep. A U-Net denoiser $\epsilon_\theta(\cdot)$ is trained to predict the injected noise:
\begin{equation}
\mathcal{L}_{\mathrm{diff}}
=
\mathbb{E}_{z_0,\epsilon,s}
\left[
\left\|
\epsilon - \epsilon_\theta(z_s, s)
\right\|_2^2
\right].
\end{equation}
At inference time, the denoising process produces an estimated clean latent $\hat{z}_0$, which is decoded back to the image space,
\begin{equation}
\hat{x} = D(\hat{z}_0),
\end{equation}
where $D(\cdot)$ denotes the VAE decoder. The anomaly heat map is obtained from the discrepancy between the input image and its restored normal counterpart,
\begin{equation}
M(x) = \psi(x,\hat{x}),
\quad
\sigma(x) = \operatorname{Pool}(M(x)),
\end{equation}
where $\psi(\cdot,\cdot)$ denotes a residual operator, $M(x)$ is the pixel-wise anomaly map, and $\sigma(x)$ is the image-level anomaly score. If the input is normal, the reconstructed image should remain close to the observation. In contrast, if the input contains anomalies, the diffusion prior tends to repair them into normal structures, making the residual map a reliable indicator of abnormality.

Since the generative prior of Stable Diffusion provides a stable reconstruction foundation, we keep the diffusion U-Net backbone frozen during continual adaptation. 
Instead of modifying the backbone directly, we attach LoRA banks to selected convolutional and linear operators, where each bank stores category-specific normality residuals. 
This design separates newly acquired normality information from the shared diffusion backbone, allowing historical normality priors to be preserved while new-category residual normality patterns are incorporated sequentially.

\subsection{History-Frozen Orthogonal LoRA Bank}

To incorporate new-category normality residuals without disturbing previously acquired normality priors, we introduce the History-Frozen Orthogonal LoRA Bank (HF-OLB). The central idea is that banks learned for previous tasks are frozen permanently, while new tasks are handled by appending new low-rank banks in directions that are explicitly orthogonal to the historical LoRA subspace.

For each adapted layer $l$, let $W_l$ denote the frozen backbone weight and let $x_l$ denote the layer input in matrix form. For convolutional layers, $x_l$ is formed by unfolding local patches, so that both convolutional and linear operators can be written consistently as matrix multiplications. We attach a layer-wise LoRA bank collection
\begin{equation}
\mathcal{L}_l^{(t)}
=
\bigcup_{q=1}^{t}
\left\{
\Delta W_{l,q,j}
\right\}_{j=1}^{m_l^{(q)}},
\end{equation}
where $t$ is the continual task index, $q$ is the task index associated with a historical or current bank, and $j$ indexes the banks appended within task $q$. Here, $m_l^{(q)}$ denotes the number of banks appended to layer $l$ during task $q$. The total number of banks available at layer $l$ after task $t$ is
\begin{equation}
K_l^{(t)} = \sum_{q=1}^{t} m_l^{(q)}.
\end{equation}
The adapted response of layer $l$ is written as
\begin{equation}
y_l
=
W_l x_l
+
\operatorname{Agg}_l
\left(
\left\{
\Delta W_{l,q,j}x_l
\mid
1 \le q \le t,\ 1 \le j \le m_l^{(q)}
\right\}
\right),
\end{equation}
where $\mathrm{Agg}_l(\cdot)$ denotes the bank aggregation operator. In our implementation, both training and inference follow a unified top-$k$ sparse bank selection principle. Each LoRA bank is associated with a learnable scalar gate, and only the top-$k$ activated banks are retained for aggregation through sparse softmax normalization. Historical banks remain available in the forward pass as routing candidates, but their LoRA factors and scalar gates are frozen after their corresponding tasks. This design allows previously learned normality banks to contribute to reconstruction without being overwritten by later updates. Each bank is parameterized in
LoRA form,
\begin{equation}
\Delta W_{l,q,j}=B_{l,q,j}\tilde A_{l,q,j}, \quad
\tilde A_{l,q,j}\in\mathbb{R}^{r_l\times d_l},\quad
B_{l,q,j}\in\mathbb{R}^{o_l\times r_l},
\end{equation}
where $\tilde A_{l,q,j}$ is equal to $A_{l,q,j}$ for ordinary LoRA and denotes the projected effective factor in the hard-orthogonal mode, $r_l$ is the LoRA rank of layer $l$, $d_l$ is the input dimension after
matrix reshaping, and $o_l$ is the output dimension. For a convolutional
layer with kernel size $h_l\times w_l$ and input channel number $c_l$, we
have $d_l=c_lh_lw_l$ after unfolding.

For final testing, we use the deployment-oriented fused form of the same top-$k$ sparse aggregation rule. Specifically, the selected LoRA banks are merged into the corresponding frozen backbone weight before evaluation. Let
$\mathcal{I}^{(t)}_l=\{(q,j)\mid 1\leq q\leq t,\;1\leq j\leq m_l^{(q)}\}$
denote all available banks at layer $l$, and let $g_{l,q,j}$ be the learned
scalar gate of bank $(q,j)$. We first select the top-$k_l$ banks according
to their gate values, where $k_l=\min(k,K_l^{(t)})$:
\begin{equation}
\begin{aligned}
\mathcal{S}^{\mathrm{fuse}}_l
&=\operatorname{TopK}_{k_l}
\left(\{g_{l,q,j}\}_{(q,j)\in\mathcal{I}^{(t)}_l}\right),\\
\bar{\omega}_{l,q,j}
&=
\begin{cases}
\dfrac{\exp(g_{l,q,j})}
{\sum\limits_{(q',j')\in\mathcal{S}^{\mathrm{fuse}}_l}
\exp(g_{l,q',j'})},
& (q,j)\in\mathcal{S}^{\mathrm{fuse}}_l,\\[2mm]
0, & \text{otherwise},
\end{cases}\\
\bar{\Delta W}^{(t)}_l
&=\alpha_l
\sum_{(q,j)\in\mathcal{I}^{(t)}_l}
\bar{\omega}_{l,q,j}\Delta W_{l,q,j},\\
\widetilde{W}^{(t)}_l
&=W_l+\bar{\Delta W}^{(t)}_l,\qquad
y^{\mathrm{test}}_l=\widetilde{W}^{(t)}_l x_l .
\end{aligned}
\end{equation}
Here, $\alpha_l$ is the LoRA scaling factor. In the hard-orthogonal mode, the LoRA-A factor used in $\Delta W_{l,q,j}$ is the projected effective factor defined by the historical subspace complement. 
At evaluation, the same fused model is used for all test samples. The top-$k$ banks are selected according to their learned scalar gates and merged into the frozen backbone weights before inference, without using task labels or category labels. Therefore, the inference pipeline preserves the same sparse bank-selection principle used during continual adaptation while avoiding dynamic routing overhead during deployment.
This design preserves the contribution of highly relevant banks while preventing less informative adapters from dominating the fused residual.

The defining principle of HF-OLB is history freezing. Assume continual task $t$ arrives after tasks $1,\dots,t-1$. We optimize only the newly appended banks for task $t$, while all historical banks remain fixed:
\begin{equation}
\begin{aligned}
\Theta^{(t)}_{\mathrm{train}}
=&
\{\bar A_{l,t,j}, B_{l,t,j}, g_{l,t,j}
\mid 1 \leq j \leq m_l^{(t)}\}_{l}
\cup \{\phi_l\}_{l\in\mathcal S},\\
&
\{\bar A_{l,q,j},B_{l,q,j},g_{l,q,j}\mid q<t\}_{l,j}
\ \text{are fixed}.
\end{aligned}
\end{equation}
Here, $\phi_l$ denotes the lightweight router parameters at layer $l$. $q<t$ denotes a historical task index, while $t$ denotes the current continual task index. Neither $q$ nor $t$ is used as a diffusion denoising timestep. Therefore, current-task optimization updates only the newly appended LoRA factors, their scalar gates, and the lightweight routers, whereas all historical LoRA factors and historical scalar gates receive no gradients. This allows previous banks to be reused in the forward computation without being overwritten by later tasks.

Freezing alone is not sufficient. If newly appended banks are initialized inside the subspace already covered by historical banks, they may compete for the same representation directions and waste capacity. To avoid this, we explicitly build an orthonormal row-space basis of the historical LoRA-$A$ matrices:
\begin{equation}
U_{l,A}^{<t}
=
\operatorname{orth}
\left(
\operatorname{rowspan}
\left\{
A_{l,q,j}
\mid
q<t,\ 1 \le j \le m_l^{(q)}
\right\}
\right),
\quad
U_{l,A}^{<t} \in \mathbb{R}^{\rho_{l,A}^{<t} \times d_l},
\end{equation}
where the rows of $U_{l,A}^{<t}$ are orthonormal and $\rho_{l,A}^{<t}$ is the dimension of the accumulated historical LoRA-$A$ subspace. The corresponding projection matrix in the input-feature space is
\begin{equation}
P_{l,A}^{<t}
=
\left(U_{l,A}^{<t}\right)^\top U_{l,A}^{<t}
\in
\mathbb{R}^{d_l \times d_l}.
\end{equation}

During optimization, we maintain an unconstrained trainable factor 
$\bar{A}_{l,t,j}$, while the effective LoRA-A factor used in each forward pass 
is projected onto the orthogonal complement of the historical LoRA-A row space:
\begin{equation}
\tilde{A}_{l,t,j}
=
\bar{A}_{l,t,j}
\left(I_{d_l}-P^{<t}_{l,A}\right)
=
\bar{A}_{l,t,j}
\left[
I_{d_l}
-
\left(U^{<t}_{l,A}\right)^{\top}U^{<t}_{l,A}
\right].
\end{equation}
Since the projection is applied to the effective factor throughout training and inference, 
$\tilde{A}_{l,t,j}(U^{<t}_{l,A})^{\top}=0$ holds at every forward pass. 
For $\Delta W_{l,t,j}=B_{l,t,j}\tilde{A}_{l,t,j}$, we have 
$\mathrm{row}(\Delta W_{l,t,j})\subseteq\mathrm{row}(\tilde{A}_{l,t,j})$; therefore, the LoRA update is constrained on the input-feature side by the same historical-subspace complement, while $B_{l,t,j}$ remains free to map residual directions to output channels. For a historical bank appended at task $q<t$, its effective factor 
$\tilde A_{l,q,j}$ is computed with the historical basis available at task $q$ 
and then kept fixed.

The benefit of HF-OLB is two-fold. 
History freezing keeps the reconstruction behavior encoded by previous banks unchanged, while the hard projection of the effective LoRA-A factor prevents newly appended banks from reusing input-side directions already occupied by historical normality bases. 
This combination preserves historical normality priors while allowing new-category residual normality information to be written into non-overlapping subspaces.

\subsection{Hierarchical Novelty-Adaptive Bank Growth}

A fixed LoRA-bank budget for all layers is rarely optimal. Different U-Net stages play different roles: high-resolution convolutional blocks are sensitive to local textures and defect boundaries, while middle attention or linear layers primarily encode more global context. Motivated by this observation, we design Hierarchical Novelty-Adaptive Bank Growth (HNABG), which combines hierarchical rank allocation with layer-wise capacity expansion.

\subsubsection{Hierarchical Rank Allocation}

Instead of assigning the same LoRA rank to all adapted layers, we allocate layer-dependent ranks according to their functional roles:
\begin{equation}
r_l =
\begin{cases}
r_{\mathrm{high}}, & l \in \mathcal{S}_{\mathrm{high}}^{\mathrm{conv}}, \\
r_{\mathrm{mid}}, & l \in \mathcal{S}_{\mathrm{mid}}^{\mathrm{attn}} \cup \mathcal{S}_{\mathrm{mid}}^{\mathrm{lin}}, \\
r_{\mathrm{base}}, & \mathrm{otherwise}.
\end{cases}
\end{equation}
Here, $\mathcal{S}_{\mathrm{high}}^{\mathrm{conv}}$ denotes high-resolution convolutional blocks in the U-Net, while $\mathcal{S}_{\mathrm{mid}}^{\mathrm{attn}}$ and $\mathcal{S}_{\mathrm{mid}}^{\mathrm{lin}}$ denote middle-block attention and linear layers, respectively. We assign a larger rank to high-resolution convolutional blocks because they must encode fine-grained textures, small structural defects, and boundary distortions. A medium rank is used for middle attention and linear layers because they require sufficient expressiveness for semantic coordination but do not need the same level of local detail modeling. The remaining adapted layers keep a base rank to avoid assigning excessive residual normality capacity to layers with limited novelty.

\subsubsection{Novelty-guided Layer-wise Capacity Growth}

Even with hierarchical ranks, not every layer requires additional banks when a new task arrives. Uniformly growing all layers may introduce redundant residual subspaces and occupy directions that should remain available for future normality patterns. Therefore, we trigger expansion only in layers whose current banks cannot adequately explain the new task.

At the beginning of continual task $t$, we compute subspace novelty statistics from sampled layer activations. For layer $l$, let
\begin{equation}
F_l^{(t)}
=
\left[
f_{l,1}^{(t)},
f_{l,2}^{(t)},
\dots,
f_{l,n_l}^{(t)}
\right]
\in
\mathbb{R}^{d_l \times n_l}
\end{equation}
be the activation matrix formed from sampled layer inputs. For linear layers, $f_{l,i}^{(t)}$ is obtained by flattening token features. For convolutional layers, local patches are obtained by unfolding, optionally after spatial pooling to reduce redundancy, and are treated as column vectors.

Given the historical LoRA-$A$ row-space basis $U_{l,A}^{<t}$ and projection matrix $P_{l,A}^{<t}$, the novelty score of task $t$ at layer $l$ is defined as
\begin{equation}
e_l^{(t)}
=
\frac{
\left\|
\left(
I_{d_l} - P_{l,A}^{<t}
\right)
F_l^{(t)}
\right\|_F^2
}{
\left\|
F_l^{(t)}
\right\|_F^2
+
\varepsilon
}.
\end{equation}
This quantity is the fraction of activation energy that lies outside the subspace already spanned by historical LoRA-$A$ directions. If $e_l^{(t)}$ is small, the current layer capacity is already sufficient, and adding a new bank would be redundant. If $e_l^{(t)}$ is large, the layer contains genuinely new directions and should receive additional capacity.

To account for layer heterogeneity, we use layer-dependent novelty thresholds:
\begin{equation}
\eta_l =
\begin{cases}
\gamma_{\mathrm{high}}\eta_0, & l \in \mathcal{S}_{\mathrm{high}}^{\mathrm{conv}}, \\
\gamma_{\mathrm{mid}}\eta_0, & l \in \mathcal{S}_{\mathrm{mid}}^{\mathrm{attn}} \cup \mathcal{S}_{\mathrm{mid}}^{\mathrm{lin}}, \\
\eta_0, & \mathrm{otherwise},
\end{cases}
\end{equation}
where $\eta_0$ is the base novelty threshold. In practice, high-resolution convolutional blocks can use a lower effective threshold because they are more sensitive to subtle appearance shifts. Middle attention and linear layers use stricter thresholds because each bank there is already relatively expressive. 
When $e_l^{(t)}$ exceeds $\eta_l$, the historical LoRA banks in layer $l$ are frozen, and new LoRA banks are appended. The number of newly appended banks is determined by the excess novelty:
\begin{equation}
m_l^{(t)}
=
\begin{cases}
0, & e_l^{(t)} \le \eta_l, \\
\max\left\{
0,
\min\left(
M_{\mathrm{task}},
M_l^{\max} - K_l^{(t-1)},
1 + \left\lfloor
\frac{e_l^{(t)} - \eta_l}{\delta}
\right\rfloor
\right)
\right\},
& e_l^{(t)} > \eta_l,
\end{cases}
\end{equation}
where $m_l^{(t)}$ is the number of banks added by task $t$ to layer $l$, $K_l$ is the current number of banks at layer $l$, $M_l^{\text{max}}$ is the layer-wise capacity limit, $M_{\text{task}}$ is the maximum number of banks allowed to grow in one task, and $\delta$ is the novelty step. This rule is preferable to adding a constant number of banks because it lets capacity scale with task difficulty: mildly novel layers receive only a small expansion, while strongly novel layers are granted more room.

To keep the growth budget focused, we expand only the top-$K_{\mathrm{grow}}$ layers ranked by novelty. Let $\mathcal{G}_t$ be the selected growth set:
\begin{equation}
\mathcal{G}_t
=
\operatorname{TopK}_{K_{\mathrm{grow}}}
\left(
\left\{
l \mid e_l^{(t)} > \eta_l
\right\},
e_l^{(t)}
\right).
\end{equation}
Here, $\operatorname{TopK}_{K_{\mathrm{grow}}}(\cdot)$ returns at most $K_{\mathrm{grow}}$ candidate layers with the largest novelty scores. If fewer than $K_{\mathrm{grow}}$ layers satisfy $e_l^{(t)}>\eta_l$, all satisfying layers are selected.

The layer-wise LoRA bank collection is then updated as
\begin{equation}
\mathcal{L}_l^{(t)}
=
\begin{cases}
\mathcal{L}_l^{(t-1)}
\cup
\left\{
\Delta W_{l,t,j}
\right\}_{j=1}^{m_l^{(t)}},
& l \in \mathcal{G}_t, \\
\mathcal{L}_l^{(t-1)},
& l \notin \mathcal{G}_t.
\end{cases}
\end{equation}
Before appending new banks, all historical banks in the selected layer are frozen. Therefore, each growth round becomes an explicit residual allocation for the new task, rather than a destructive update to previously learned banks.

HNABG offers two advantages for normality-prior preservation. 
First, it avoids redundant bank growth in layers whose historical subspaces already explain the incoming normal patterns, thereby reserving residual subspace directions for future categories. 
Second, by expanding only where novelty is high, it reduces unnecessary overlap among normality representations and prevents the LoRA bank space from becoming fragmented. 

Overall, HF-OLB and HNABG are complementary. 
HF-OLB prevents newly added LoRA banks from overwriting historical normality priors, while HNABG determines where and how much residual normality capacity should be introduced. 
Together, they enable continual U-Net adaptation through a controlled process of normality-preserving residual subspace expansion. 
Algorithm~\ref{alg:pead} summarizes the overall continual training and fused inference pipeline of our framework.
\begin{algorithm}[t]
\caption{Continual training and fused inference.}
\label{alg:pead}
\footnotesize
\begin{algorithmic}[1]
\Require Normal task stream $\{\mathcal{D}^{\mathrm{tr}}_t\}_{t=1}^{T}$;
frozen diffusion backbone $(E,D,\epsilon_{\theta})$;
adapted layer set $\mathcal{S}$; top-$k$ value; bank-growth configuration.
\Ensure Fused model $\{W^{f(T)}_l\}_{l\in\mathcal{S}}$,
anomaly map $M(I)$, and image-level score $\sigma(I)$.

\State Freeze $(E,D,\epsilon_{\theta})$ and initialize
$\mathcal{L}^{(0)}_l=\varnothing,\ \forall l\in\mathcal{S}$.

\Statex \hspace*{-\algorithmicindent}\textbf{-- Continual training}
\For{$t=1$ to $T$}
    \State Extract $\{x_l,F_l^{(t)}\}_{l\in\mathcal{S}}$
    from $\mathcal{D}^{\mathrm{tr}}_t$ with the frozen U-Net.
    
    \ForAll{$l\in\mathcal{S}$}
        \State Build $U^{<t}_{l,A}$ and $P^{<t}_{l,A}$ by Eqs.~(12)--(13).
        \State Compute novelty score $e_l^{(t)}$ by Eq.~(17).
    \EndFor

    \State Compute $\{m_l^{(t)}\}$ by Eq.~(19) and select $\mathcal{G}_t$ by Eq.~(20).

    \State Freeze $\mathcal{L}^{(t-1)}_l$ and append
    $\{\Delta W_{l,t,j}\}_{j=1}^{m_l^{(t)}}$ for each $l\in\mathcal{G}_t$.

    \State Project each new effective LoRA-A factor by Eq.~(14).

    \State Train $\Theta_{\mathrm{train}}^{(t)}$ under Eq.~(11) with $\mathcal{L}_{\mathrm{diff}}$ in Eq.~(3).
\EndFor

\Statex \hspace*{-\algorithmicindent}\textbf{-- Fused inference}
\ForAll{$l\in\mathcal{S}$}
    \State Select top-$k$ LoRA banks and fuse them into $W^{f(T)}_l$ by Eq. (10).
\EndFor

\State Reconstruct $\hat{I}=D(\hat{z}_0)$ and compute
$M(I)$ and $\sigma(I)$ by Eqs.~(4)--(5).

\State \Return $\{W^{f(T)}_l\}_{l\in\mathcal{S}}$, $M(I)$, and $\sigma(I)$.
\end{algorithmic}
\end{algorithm}

\section{Experiments and Analysis}

\subsection{Experimental Setup}

\noindent \textbf{Datasets.} The datasets used in this paper are MVTec~\cite{bergmann2019mvtec} and VisA~\cite{zou2022spot}. Both datasets contain multiple categories. Specifically, MVTec contains 15 categories, with image resolutions ranging from $700^2$ to $900^2$ pixels. It includes 3629 training samples, all of which are normal, and 1725 test samples, including 467 normal samples and 1258 anomalous samples. VisA contains 12 categories, with image resolutions of roughly $1.5K\times1K$ pixels. It includes 8661 training samples, all of which are normal, and 2162 test samples, including 962 normal samples and 1200 anomalous samples. These two datasets contain diverse industrial objects and textures and are widely used to evaluate real-world anomaly detection methods.
\begin{table*}[t]
\centering
\caption{LoRA insertion blocks and rank allocation in the Stable-Diffusion U-Net. The three row groups separated by horizontal rules correspond to the input path, middle block, and output path, respectively. Curly braces denote grouped blocks or submodules that share the same LoRA rank.}
\label{tab:lora_block_rank}
\scriptsize
\setlength{\tabcolsep}{5pt}
\renewcommand{\arraystretch}{1.08}
\begin{tabular}{
p{0.15\textwidth}
p{0.37\textwidth}
p{0.43\textwidth}
}
\toprule
\textbf{Rank} &
\textbf{Block name} &
\textbf{LoRA-inserted submodules} \\
\midrule

\(r_{\mathrm{high}}=16\) &
\texttt{input\_blocks.0} &
\texttt{0} \\

\(r_{\mathrm{high}}=16\) &
\texttt{input\_blocks.\{1,2,4,5,7,8,10,11\}} &
\texttt{0.\{in\_layers.2,out\_layers.3\}} \\

\(r_{\mathrm{high}}=16\) &
\texttt{input\_blocks.\{4,7\}} &
\texttt{0.skip\_connection} \\

\(r_{\mathrm{high}}=16\) &
\texttt{input\_blocks.\{3,6,9\}} &
\texttt{0.op} \\

\(r_{\mathrm{high}}=16\) &
\texttt{input\_blocks.\{1,2,4,5,7,8\}} &
\texttt{1.\{proj\_in,proj\_out\}} \\

\(r_{\mathrm{mid}}=8\) &
\texttt{input\_blocks.\{1,2,4,5,7,8\}} &
\texttt{1.attn\{1,2\}.to\_\{q,k,v,out.0\}} \\

\(r_{\mathrm{base}}=4\) &
\texttt{input\_blocks.\{1,2,4,5,7,8\}} &
\texttt{1.ff.net.\{0.proj,2\}} \\

\midrule

\(r_{\mathrm{base}}=4\) &
\texttt{middle\_block.\{0,2\}} &
\texttt{\{in\_layers.2,out\_layers.3\}} \\

\(r_{\mathrm{base}}=4\) &
\texttt{middle\_block.1} &
\texttt{\{proj\_in,proj\_out\}} \\

\(r_{\mathrm{mid}}=8\) &
\texttt{middle\_block.1} &
\texttt{attn\{1,2\}.to\_\{q,k,v,out.0\}; ff.net.\{0.proj,2\}} \\

\midrule

\(r_{\mathrm{high}}=16\) &
\texttt{output\_blocks.\{0,\ldots,11\}} &
\texttt{0.\{in\_layers.2,out\_layers.3,skip\_connection\}} \\

\(r_{\mathrm{high}}=16\) &
\texttt{output\_blocks.2} &
\texttt{1.conv} \\

\(r_{\mathrm{high}}=16\) &
\texttt{output\_blocks.\{5,8\}} &
\texttt{2.conv} \\

\(r_{\mathrm{high}}=16\) &
\texttt{output\_blocks.\{3,\ldots,11\}} &
\texttt{1.\{proj\_in,proj\_out\}} \\

\(r_{\mathrm{mid}}=8\) &
\texttt{output\_blocks.\{3,\ldots,11\}} &
\texttt{1.attn\{1,2\}.to\_\{q,k,v,out.0\}} \\

\(r_{\mathrm{base}}=4\) &
\texttt{output\_blocks.\{3,\ldots,11\}} &
\texttt{1.ff.net.\{0.proj,2\}} \\

\bottomrule
\end{tabular}
\end{table*}
\noindent\textbf{Task Settings.} We evaluate our method on both MVTec and VisA under multiple incremental settings. For MVTec, the task streams include 14--1 (1 step), $3\times5$ (5 steps), and 7--$1\times8$ (8 steps). For VisA, we use 11--1 (1 step), 8--$2\times2$ (2 steps), and $2\times6$ (6 steps). Following the one-for-more continual IAD protocol, we evaluate each setting under three randomly shuffled task orders. 
For each random order, all compared methods use the same category sequence, data split, and training configuration to ensure a fair comparison. 
The reported A-AUROC and FM results are averaged over the three task-order trials. 
To further evaluate long-horizon stability, we include extended incremental settings with more sequential steps.

\noindent\textbf{Evaluation Metrics.} 
To evaluate model performance, we calculate the average AUROC (A-AUROC) and the forgetting measure (FM) over $N$ continual steps~\cite{li2025one}. 
Both A-AUROC and FM are reported in the image-level/pixel-level format. 
For each continual setting, we use three randomly shuffled task orders, where all compared methods share the same order within each trial. 
The reported A-AUROC and FM values are averaged over the three trials.

\noindent\textbf{Implementation Details.} We use a 4090 GPU for training. We adopt a Stable Diffusion v1.5-based latent diffusion backbone with a 4-level U-Net, where the channel multiplier is $[1,2,4,4]$, each level contains 2 ResBlocks, the transformer depth is 1, the number of attention heads is 8, and the base channel number is 320. The backbone also uses an AutoencoderKL with latent dimension $z=4$ and 1000 diffusion steps, and we train the model using the AdamW optimizer with an initial learning rate of $1 \times 10^{-4}$. In all experiments and comparisons, we use the same experimental configuration: input images are resized to $256\times256$, the maximum number of epochs is 300, the batch size is 4, and a PyTorch DataLoader with \texttt{num\_workers=4} is used for data loading.
\begin{table*}[t]
\centering
\small
\caption{Image-level/pixel-level results of our method on MVTec under 3 continual anomaly detection settings, reported with A-AUROC and FM. The best and second-best results are marked in \textbf{bold} and \underline{underlined}. $\dagger$ indicates diffusion-based continual IAD methods, and * indicates LoRA-based continual diffusion adaptation variants. \textcolor{ForestGreen}{Green} denotes methods based on official public code, whereas \textcolor{NavyBlue}{Blue} denotes methods implemented and tuned by our team.}
\label{tab:mvtec}
\resizebox{\textwidth}{!}{%
\begin{tabular}{l l|cc|cc|cc}
\hline
\multirow{2}{*}{Method} 
& \multirow{2}{*}{Venue}
& \multicolumn{2}{c|}{14--1 with 1 Step}
& \multicolumn{2}{c|}{3 $\times$ 5 with 5 Steps}
& \multicolumn{2}{c}{7--1 $\times$ 8 with 8 Steps} \\
\cline{3-8}
& 
& A-AUROC ($\uparrow$) & FM ($\downarrow$)
& A-AUROC ($\uparrow$) & FM ($\downarrow$)
& A-AUROC ($\uparrow$) & FM ($\downarrow$) \\
\hline

\textcolor{ForestGreen}{UCAD}\cite{liu2024unsupervised} & AAAI 24
& 92.1/94.4 & 2.3/1.9
& 81.8/87.2 & 10.4/9.5
& 75.6/79.9 & 12.1/9.1
\\

\textcolor{ForestGreen}{MambaAD}\cite{he2024mambaad} & NeurIPS 24
& 80.5/-- & 10.8/--
& 75.8/-- & 15.6/--
& 72.3/-- & 21.4/--
\\

\textcolor{ForestGreen}{IUF}\cite{tang2024incremental} & ECCV 24
& 94.2/95.5 & 1.8/1.5
& 83.9/87.7 & 9.7/8.4
& 76.8/81.0 & 10.2/9.5
\\

\cline{1-8}

\textcolor{NavyBlue}{ReplayCAD}\cite{hu2025replaycad}$\dagger$ & IJCAI 25
& \textbf{95.1}/\underline{95.3} & 0.2/\underline{0.1}
& \underline{87.0}/88.1 & 6.1/4.7
& 83.3/\underline{86.1} & 8.5/\underline{3.8}
\\

\textcolor{ForestGreen}{CDAD}\cite{li2025one}$^\dagger$ & CVPR 25
& \underline{94.9}/\textbf{95.6} & \textbf{-0.2}/\textbf{-0.1}
& 88.5/\underline{91.7} & \underline{4.7}/\underline{3.5}
& \underline{83.5}/85.6 & \underline{6.0}/4.3
\\

\cline{1-8}

\textcolor{NavyBlue}{Seq-SD-LoRA}\cite{hu2021lora}* & ICLR 22
& 85.8/86.8 & 4.7/4.3
& 71.5/79.5 & 13.4/10.0
& 68.1/78.2 & 19.7/15.2
\\

\textcolor{NavyBlue}{Seq-SD-CLLoRA}\cite{he2025cllora}* & CVPR 25
& 89.5/90.4 & 2.9/2.6
& 82.5/85.6 & 5.7/5.2
& 77.1/80.7 & 6.5/8.0
\\

\cline{1-8}
\rowcolor{gray!15}
\textbf{Ours}* & -
& 93.0/93.8 & \underline{0.1}/\underline{0.1}
& \textbf{86.7}/\textbf{92.4} & \textbf{3.3}/\textbf{3.1}
& \textbf{84.1}/\textbf{91.5} & \textbf{3.8}/\textbf{3.2}
\\
\hline
\end{tabular}}
\end{table*}

For reproducibility, Table~\ref{tab:lora_block_rank} summarizes the exact LoRA insertion positions and the corresponding rank allocation in the Stable-Diffusion U-Net. For sparse routing and fused inference, we set k=4 for all experiments, following the same configuration across datasets and task streams. In our implementation, LoRA adapters are attached only to selected Conv2d and Linear operators, including convolutional layers, projection layers, attention-related linear layers, and feed-forward layers. The rank is assigned according to the U-Net region and operator type: high-resolution convolutional layers are given a larger rank of \(r_{\mathrm{high}}=16\) to enhance adaptation capacity, middle attention/linear layers use \(r_{\mathrm{mid}}=8\), and the remaining adapted layers use \(r_{\mathrm{base}}=4\). For novelty filtering, we set a base threshold \(\tau_0=0.2\) and apply group-wise scaling factors following the same rank hierarchy, where \(\gamma_{\mathrm{high}}=0.8\), \(\gamma_{\mathrm{mid}}=1.2\), and \(\gamma_{\mathrm{base}}=1.0\). Accordingly, the thresholds are computed as 0.16, 0.24, and 0.20.

\subsection{Comparison Results}

\begin{table*}[t]
\centering
\small
\caption{Image-level/pixel-level results of our method on VisA under 3 continual anomaly detection settings, reported with A-AUROC and FM. The best and second-best results are marked in \textbf{bold} and \underline{underlined}. $\dagger$ indicates diffusion-based continual IAD methods, and * indicates LoRA-based continual diffusion adaptation variants. \textcolor{ForestGreen}{Green} denotes methods based on official public code, whereas \textcolor{NavyBlue}{Blue} denotes methods implemented and tuned by our team.}
\label{tab:visa}
\resizebox{\textwidth}{!}{%
\begin{tabular}{l l|cc|cc|cc}
\hline
\multirow{2}{*}{Method} 
& \multirow{2}{*}{Venue}
& \multicolumn{2}{c|}{11--$1\times1$ with 1 Step}
& \multicolumn{2}{c|}{8--$2\times2$ with 2 Steps}
& \multicolumn{2}{c}{2 $\times$ 6 with 6 Steps} \\
\cline{3-8}
&
& A-AUROC ($\uparrow$) & FM ($\downarrow$)
& A-AUROC ($\uparrow$) & FM ($\downarrow$)
& A-AUROC ($\uparrow$) & FM ($\downarrow$) \\
\hline

\textcolor{ForestGreen}{UCAD}\cite{liu2024unsupervised} & AAAI 24
& 84.2/\underline{95.6} & 3.0/0.9
& 80.7/92.5 & 5.2/5.8
& 73.4/76.8 & 15.0/10.1
\\

\textcolor{ForestGreen}{MambaAD}\cite{he2024mambaad} & NeurIPS 24
& 78.0/-- & 13.2/--
& 73.9/-- & 20.2/--
& 64.7/-- & 33.8/--
\\

\textcolor{ForestGreen}{IUF}\cite{tang2024incremental} & ECCV 24
& 84.6/\textbf{95.7} & 2.7/1.4 
& 81.3/91.8 & 5.8/6.5 
& 78.9/80.4 & 14.2/11.1
\\

\cline{1-8}

\textcolor{NavyBlue}{ReplayCAD}\cite{hu2025replaycad}$\dagger$ & IJCAI 25
& \underline{90.9}/94.6 & 0.7/0.6 
& 84.5/92.9 & 3.0/\textbf{1.2}
& 79.8/85.1 & 9.8/7.9
\\

\textcolor{ForestGreen}{CDAD}\cite{li2025one}$^\dagger$ & CVPR 25
& \textbf{93.7}/95.5 & \textbf{0.3}/\textbf{0.1} 
& \textbf{85.9}/\textbf{93.2} & \textbf{1.4}/\underline{1.3}
& \underline{80.8}/\underline{88.6} & \underline{6.2}/5.2
\\

\cline{1-8}

\textcolor{NavyBlue}{Seq-SD-LoRA}\cite{hu2021lora}* & ICLR 22
& 84.2/86.9 & 2.1/1.9 
& 80.7/81.0 & 5.0/4.6 
& 68.0/70.4 & 18.7/16.3
\\

\textcolor{NavyBlue}{Seq-SD-CLLoRA}\cite{he2025cllora}* & CVPR 25
& 86.9/90.7 & 0.7/0.7 
& 83.0/83.2 & 3.1/2.8 
& 73.7/78.6 & 13.6/9.5
\\

\cline{1-8}
\rowcolor{gray!15}
\textbf{Ours}* & -
& 89.6/93.3 & \underline{0.5}/\underline{0.3}
& \underline{85.2}/\underline{93.0} & \underline{1.7}/\underline{1.3}
& \textbf{83.6}/\textbf{91.8} & \textbf{3.8/3.9}
\\
\hline
\end{tabular}}
\end{table*}

The compared methods can be divided into three groups: 
(1) conventional continual anomaly detection methods, including UCAD~\cite{liu2024unsupervised}, MambaAD~\cite{he2024mambaad}, and IUF~\cite{tang2024incremental}; 
(2) diffusion-based continual IAD methods, including ReplayCAD~\cite{hu2025replaycad} and CDAD~\cite{li2025one}; 
and (3) LoRA-based continual diffusion adaptation variants, including Seq-SD-LoRA~\cite{hu2021lora} and Seq-SD-CLLoRA~\cite{he2025cllora}. 
In Seq-SD-LoRA, each incoming task is assigned an independent LoRA unit on the Stable Diffusion backbone, so that task-specific residual updates are stored in isolated adapters without explicit historical-subspace protection. 
In Seq-SD-CLLoRA, each adapted layer contains one shared LoRA branch initialized from the first task to retain common normality features, while the remaining LoRA branches store task-specific residual features beyond the shared component. 
The third group therefore evaluates whether the proposed history-frozen orthogonal bank and novelty-adaptive growth designs can better preserve historical normality priors and allocate residual normality capacity than isolated LoRA units or shared-specific LoRA organization under the same diffusion reconstruction framework.

As shown in Tables~\ref{tab:mvtec} and~\ref{tab:visa}, diffusion-based continual IAD methods can still achieve strong results in short task streams, where historical normality-prior drift is relatively limited. 
However, our method shows clearer advantages over the LoRA-based continual diffusion adaptation variants. 
Compared with Seq-SD-LoRA and Seq-SD-CLLoRA, our method consistently achieves higher A-AUROC and lower FM on both datasets, indicating that isolated LoRA units or shared-specific LoRA organization are insufficient to fully protect historical normality priors. 

The advantage becomes more evident as the task stream grows. 
On MVTec $3 \times 5$, our method achieves the best pixel-level A-AUROC (92.4) and the lowest FM (3.1). 
On the longest MVTec setting, $7$--$1 \times 8$, it ranks first on all four metrics, reaching 84.1/91.5 A-AUROC with 3.8/3.2 FM. 
A similar trend appears on VisA. 
In the hardest setting, $2 \times 6$, our method obtains 83.6/91.8 A-AUROC and 3.8/3.9 FM, improving the second-best A-AUROC by 2.8/3.2 points and reducing FM by 2.4/1.3. 
These results show that the proposed normality-preserving subspace expansion is especially effective in long-horizon continual anomaly detection.

\subsection{Ablation Study}
\begin{table}[t]
\centering
\caption{Ablation study of key design choices on MVTec under setting 2. 
``Freeze'' denotes freezing historical LoRA banks, and ``A-Proj.'' denotes hard projection of the effective LoRA-A factor onto the complement of the historical LoRA-A subspace. 
For ``Growth Strategy'', ``None'' means no additional bank growth, ``Fixed'' means uniformly expanding all LoRA-inserted layers once growth is triggered, and ``Adaptive'' means novelty-guided layer-wise growth. 
The best results are marked in \textbf{bold}.}
\label{tab:ablation}
\renewcommand{\arraystretch}{1.12}
\setlength{\tabcolsep}{4.2pt}
\resizebox{\linewidth}{!}{
\begin{tabular}{c|l|c|c|c|c|c}
\hline
No. & Variant 
& Freeze 
& A-Proj. 
& Growth Strategy 
& A-AUROC ($\uparrow$)
& FM ($\downarrow$) \\
\hline

1 & Standard LoRA 
& \xmark & \xmark & None 
& 71.5/79.5 & 13.4/10.0 \\
\hline

2 & HF-OLB only 
& \cmark & \cmark & None 
& 81.2/86.9 & 7.1/4.6 \\
\hline

3 & LoRA + HNABG 
& \xmark & \xmark & Adaptive 
& 79.2/82.3 & 10.6/7.5 \\
\hline

4 & Ours w/o Hard A-Proj. 
& \cmark & \xmark & Adaptive
& 83.0/87.6 & 5.8/5.2 \\
\hline

5 & Ours w/ Fixed Growth 
& \cmark & \cmark & Fixed 
& 83.8/89.6 & 4.7/4.4 \\
\hline

\rowcolor{gray!15}
6 & Ours 
& \cmark & \cmark & Adaptive
& \textbf{86.7/92.4} 
& \textbf{3.3/3.1} \\
\hline

\end{tabular}}
\end{table}
Table~\ref{tab:ablation} shows that the improvement does not come from simply inserting LoRA, but from how the LoRA banks are preserved and expanded during continual adaptation. 
Standard LoRA suffers from severe forgetting, with only 71.5/79.5 A-AUROC and 13.4/10.0 FM. 
Adding HF-OLB improves the result to 81.2/86.9 A-AUROC and reduces FM to 7.1/4.6, confirming that freezing historical banks and constraining new updates to the historical-complement subspace help preserve historical normality priors. 
LoRA + HNABG also improves over Standard LoRA, but still shows larger forgetting than HF-OLB, indicating that adaptive growth alone is insufficient without explicit history protection.
\begin{figure}[t]
\centering
\includegraphics[width=1\linewidth, trim=0 50 0 0, clip]{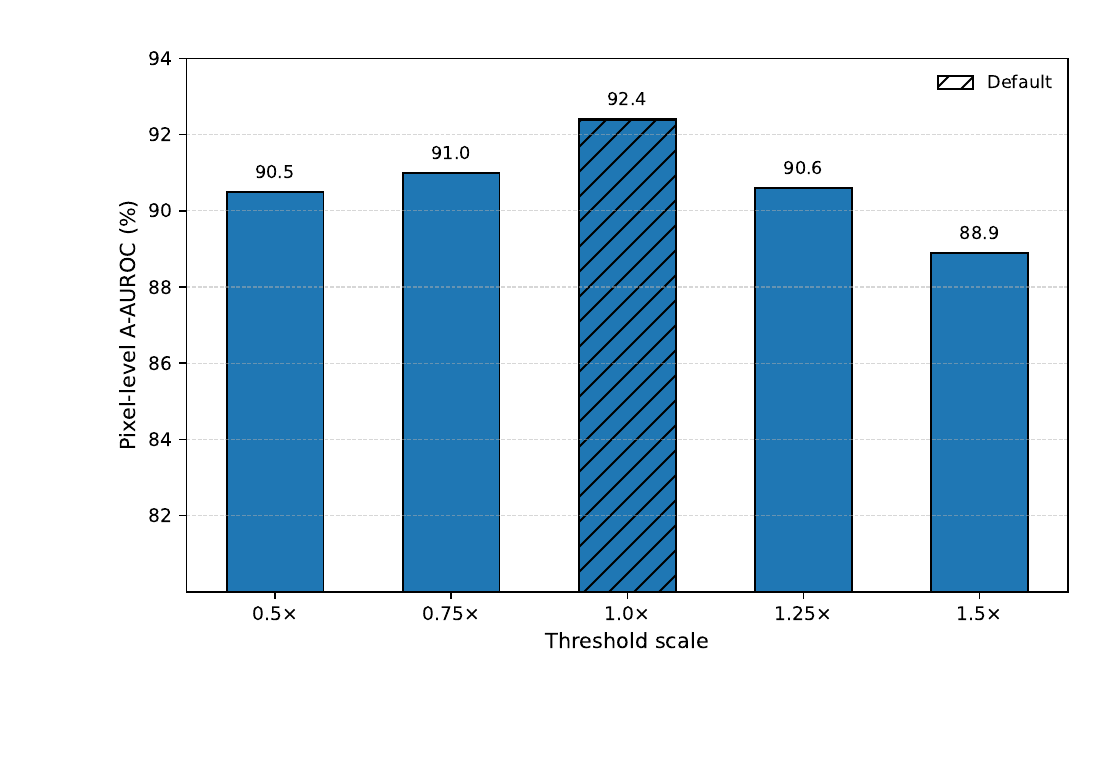}
\vspace{-2mm}
\caption{Sensitivity analysis of the three-level HNABG threshold configuration on MVTec setting 2. 
The default High/Mid/Base threshold configuration is 0.16/0.24/0.20 and is scaled by different factors while preserving their relative ratios. 
The best performance at the default scale suggests that balanced novelty-guided growth is important for allocating residual normality capacity.}
\label{fig:threshold}
\vspace{-2mm}
\end{figure}
Rows 4 and 5 further isolate the two key designs in the full framework. 
Removing hard A-projection degrades A-AUROC from 86.7/92.4 to 83.0/87.6 and increases FM from 3.3/3.1 to 5.8/5.2, showing that freezing historical banks alone cannot fully suppress subspace interference. 
Replacing adaptive growth with fixed growth also weakens performance, reducing A-AUROC to 83.8/89.6 and increasing FM to 4.7/4.4. 
This suggests that uniformly expanding LoRA banks cannot allocate residual normality capacity as effectively as novelty-guided layer-wise growth. 
The full model achieves the best results on both A-AUROC and FM, demonstrating the complementarity of HF-OLB and HNABG.

\subsection{Hyperparameter Analysis}

\begin{figure}[t]
    \centering
    \includegraphics[width=1\linewidth]{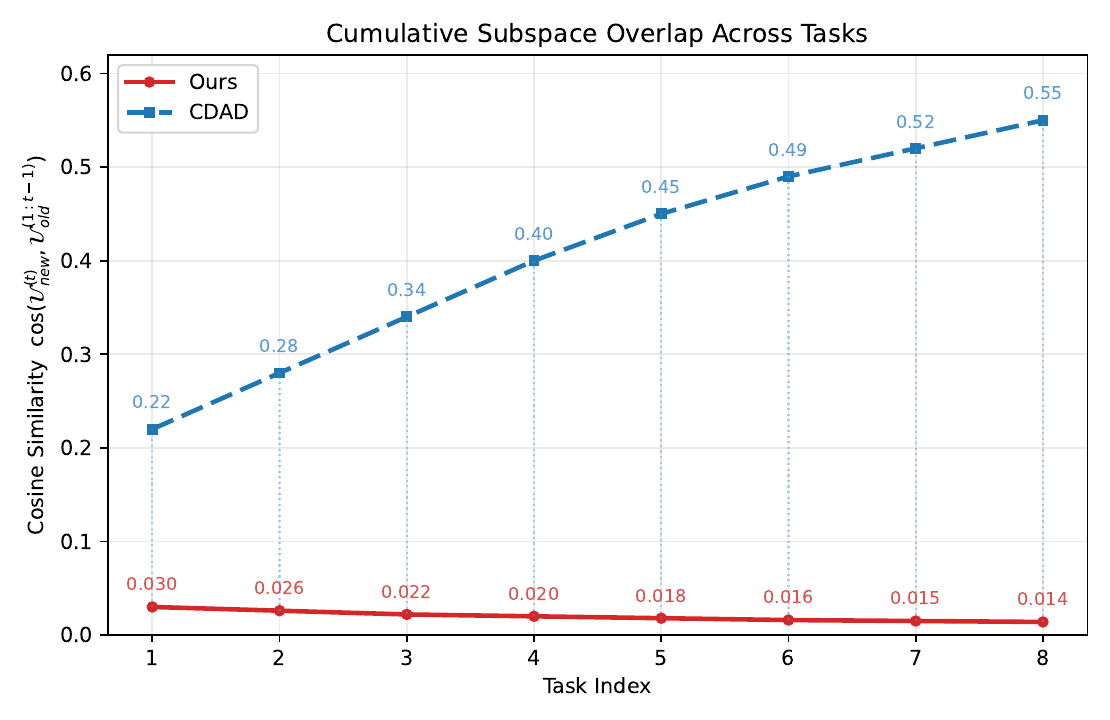}
    \caption{Comparison of cosine similarity between the newly acquired task subspace and the cumulative old-knowledge subspace on MVTec Setting 3.}
    \label{fig:cosine}
\end{figure}
Since HNABG uses a three-level threshold configuration, we analyze its sensitivity by scaling the default thresholds, i.e., 0.16/0.24/0.20, with different factors. 
As shown in Figure~\ref{fig:threshold}, the default scale achieves the best pixel-level A-AUROC of 92.4. 
When the thresholds are scaled down to 0.5$\times$ and 0.75$\times$, the performance decreases to 90.5 and 91.0, respectively, suggesting that overly loose growth criteria may introduce redundant LoRA banks. 
When the thresholds are scaled up to 1.25$\times$ and 1.5$\times$, the performance further drops to 90.6 and 88.9, indicating that overly strict growth criteria may under-allocate residual normality capacity for newly arriving categories. 
These results show that the default threshold configuration provides a better balance between preserving historical normality priors and adapting to new normality patterns.

\subsection{Effectiveness of Hard Orthogonal Projection}

As shown in Figure~\ref{fig:cosine}, we measure the cosine similarity between the newly acquired task subspace $U^{(t)}_{\mathrm{new}}$ and the cumulative old-knowledge subspace $U^{(1:t-1)}_{\mathrm{old}}$. 
As a representative gradient-constrained continual diffusion baseline, CDAD stabilizes sequential updates through soft update constraints, but it does not explicitly enforce a hard orthogonal complement for new-task adaptation. \begin{figure}[t]
    \centering
    \includegraphics[width=1\textwidth]{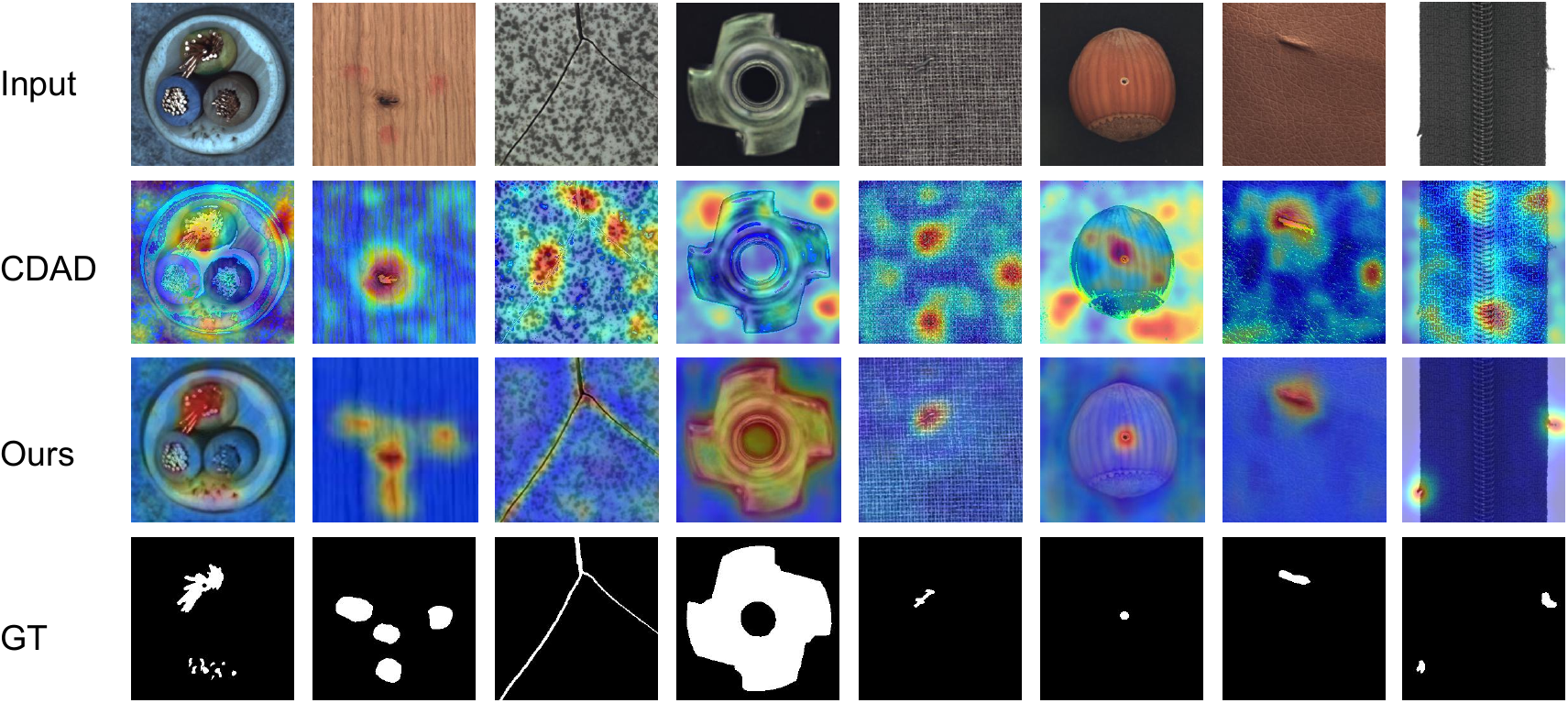}
    \caption{Qualitative comparison of anomaly localization maps produced by CDAD and the proposed method on representative industrial inspection samples. From top to bottom, each row shows the input image, the anomaly map generated by CDAD, the anomaly map generated by our method, and the ground-truth mask. Warmer colors indicate higher anomaly responses.}
    \label{fig:cdadours}
\end{figure}

Therefore, its overlap with the cumulative old subspace tends to increase as tasks accumulate, which may lead to weight drift and degraded long-horizon performance. 
In contrast, our hard-orthogonal LoRA bank projects the effective LoRA-A factor onto the orthogonal complement of previous banks at each forward pass, keeping the input-side subspace overlap near zero and thereby better preserving old knowledge.

\subsection{Visualization Results}

\begin{figure}[t]
\centering
\includegraphics[width=0.95\linewidth]{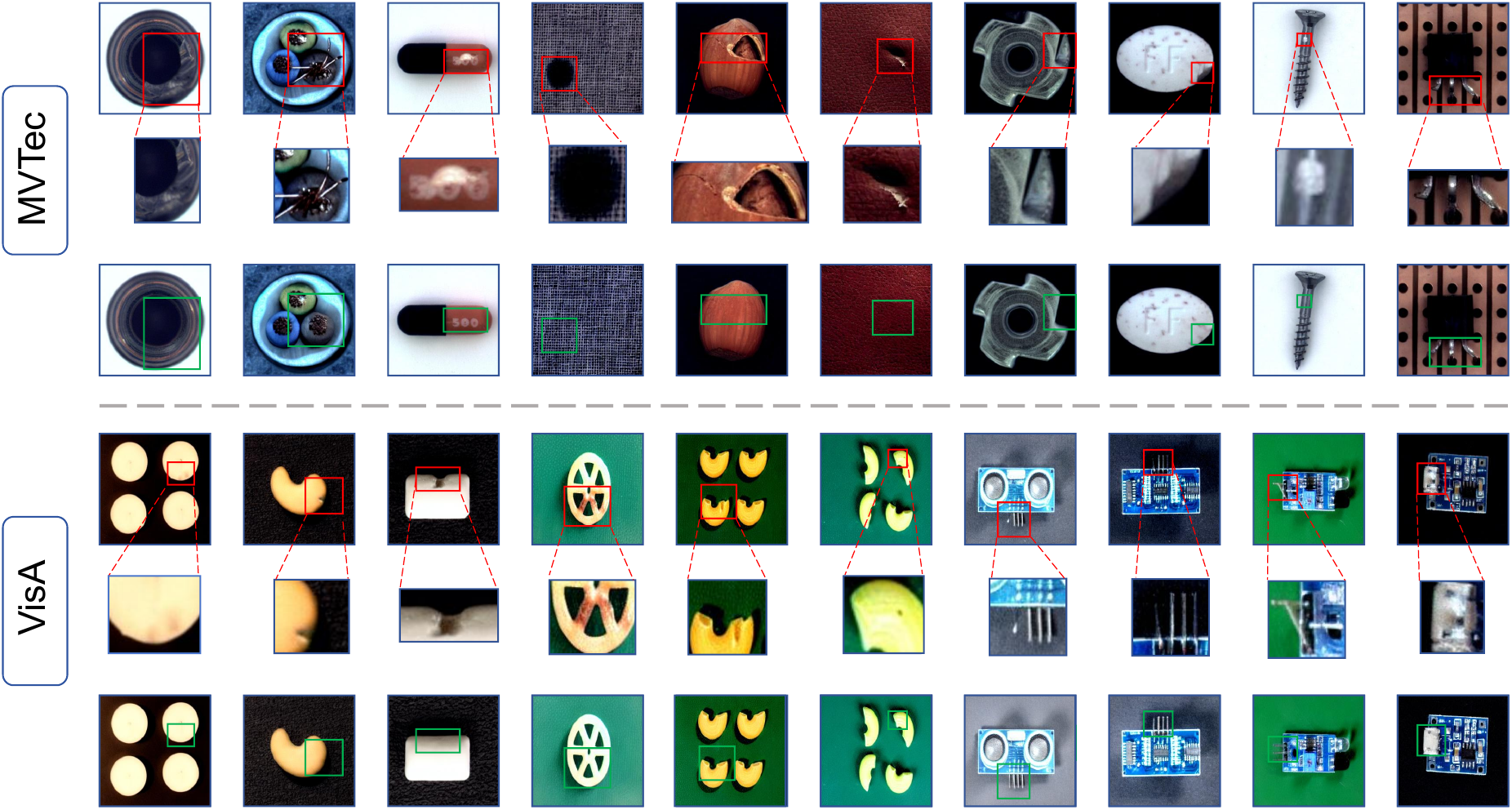}
\vspace{-2mm}
\caption{Visualization of anomaly localization and restoration on MVTec and VisA. 
For each sample, we show the anomalous image, zoomed anomalous region, and reconstructed normal image. 
\textcolor{red}{Red} and \textcolor{green}{green} boxes indicate anomalous and reconstructed normal regions, respectively.}
\label{fig:visualization}
\vspace{-3mm}
\end{figure}
As shown in Figure~\ref{fig:cdadours}, CDAD tends to produce over-diffused anomaly responses and frequently activates normal regions with complex textures or strong structural edges, which may lead to false-positive detections and inaccurate defect boundaries. In contrast, the proposed method concentrates the anomaly responses on the true defective areas and exhibits stronger consistency with the ground-truth masks. This advantage is particularly evident for subtle defects and texture-like anomalies, where the abnormal patterns are easily confused with normal background variations. The qualitative results indicate that our method not only improves the discriminability between normal and abnormal regions, but also provides more reliable pixel-level localization, thereby enhancing the interpretability and practical applicability of industrial anomaly inspection.

As shown in Figure~\ref{fig:visualization}, our method accurately localizes anomalous regions across different object and texture categories on both MVTec and VisA. The zoomed regions further demonstrate that the detected anomalies are consistent with the actual defective areas. Moreover, the reconstructed normal images effectively remove abnormal patterns while preserving the overall structure and appearance of the original samples. These qualitative results indicate that our method achieves reliable anomaly localization and normal reconstruction under diverse visual conditions.
\begin{table*}[t]
\centering
\small
\caption{Auxiliary resource analysis on MVTec Setting 1. 
$\dagger$ indicates diffusion-based continual IAD methods, and * indicates LoRA-based continual diffusion adaptation variants. 
\textcolor{ForestGreen}{Green} and \textcolor{NavyBlue}{Blue} denote methods based on official public code and our implementation, respectively.}
\label{tab:gpu}
\resizebox{\textwidth}{!}{%
\begin{tabular}{l l|c|c|c|c|c}
\hline
Method
& Venue
& Backbone Parameter
& TP ($\downarrow$)
& M-TP ($\downarrow$)
& M-TPR ($\downarrow$)
& P-GPU ($\downarrow$)
\\
\hline

\textcolor{NavyBlue}{ReplayCAD}\cite{hu2025replaycad}$\dagger$
& IJCAI 25
& 95,631,088
& 70,768,560
& 70,768,560
& 74.0\%
& 23880 MiB
\\

\textcolor{ForestGreen}{CDAD}\cite{li2025one}$^\dagger$
& CVPR 25
& 1,099,384,995
& 892,670,652
& 892,670,652
& 81.2\%
& 22896 MiB
\\

\cline{1-7}

\textcolor{NavyBlue}{Seq-SD-LoRA}\cite{hu2021lora}*
& ICLR 22
& 1,131,225,573
& 330,145,926
& 330,145,926
& 29.2\%
& 17648 MiB
\\

\textcolor{NavyBlue}{Seq-SD-CLLoRA}\cite{he2025cllora}*
& CVPR 25
& 1,131,225,573
& 531,375,210
& 351,809,684
& 31.1\%
& 18192 MiB
\\

\cline{1-7}
\rowcolor{gray!15}
\textbf{Ours}*
& -
& 1,131,225,573
& 65,345,721
& 41,482,234
& 3.6\%
& 13936 MiB
\\
\hline
\end{tabular}}
\end{table*}
\subsection{Auxiliary Resource Analysis}

Table~\ref{tab:gpu} reports the auxiliary resource usage on MVTec Setting 1. 
Here, TP denotes the number of trainable parameters, M-TP denotes the maximum number of training parameters during the continual training process, M-TPR denotes the corresponding maximum training-parameter ratio, and P-GPU denotes the peak GPU memory usage over the entire training process. 

Although resource efficiency is not the main objective of our framework, the results show that the proposed normality-preserving bank design remains practical. 
Compared with diffusion-based continual IAD methods, our method reduces the peak trainable parameters from 70.77M in ReplayCAD and 892.67M in CDAD to 41.48M. 
The corresponding M-TPR is only 3.6\%, much lower than ReplayCAD, CDAD, Seq-SD-LoRA, and Seq-SD-CLLoRA. 
In terms of memory usage, our method requires 13,936 MiB peak GPU memory, saving 9,944 MiB compared with ReplayCAD and 8,960 MiB compared with CDAD. 
Compared with LoRA-based continual diffusion variants, our method also reduces peak GPU memory by 3,712 MiB over Seq-SD-LoRA and 4,256 MiB over Seq-SD-CLLoRA. 
These results indicate that HF-OLB and HNABG preserve historical normality priors with limited additional resource overhead, rather than relying on a larger trainable parameter budget.

\FloatBarrier
\section{Conclusion}

We present a normality-preserving continual industrial anomaly detection framework built on a frozen Stable Diffusion backbone. 
The core objective is to prevent newly arriving categories from perturbing historical normality priors while still allowing the model to acquire residual normality capacity for new categories. 
To this end, HF-OLB freezes historical LoRA banks and constrains new updates to the orthogonal complement of previous normality subspaces, while HNABG adaptively determines where additional residual banks should be introduced according to layer-wise novelty. 
Together, they formulate continual diffusion adaptation as a controlled process of normality-preserving residual subspace expansion. 
On the challenging long-horizon MVTec $7$--$1\times8$ setting, our framework achieves 84.1/91.5 A-AUROC with 3.8/3.2 FM. 
On the hardest VisA $2\times6$ setting, it further reaches 83.6/91.8 A-AUROC with 3.8/3.9 FM, showing stronger stability under longer category sequences. 
These results demonstrate that historical normality priors can be effectively preserved in continual diffusion-based anomaly detection. 
Future work will extend this framework to more complex industrial scenarios, including stronger domain shifts, longer open-ended task streams, and adaptive bank selection under online inspection conditions.

\section*{Acknowledgements}

This work was supported by the National Key Research and Development Program of China under Grant No. 2022YFB3705504. The authors acknowledge the support from the Engineering Research Center of Intelligent Control Systems and Intelligent Equipment, Ministry of Education, Yanshan University.

\bibliographystyle{elsarticle-num-names} 
\bibliography{IAD}

\end{document}